\pdfoutput=1

\documentclass[11pt]{article}

\usepackage[]{ACL2023}

\usepackage{times}
\usepackage{latexsym}

\usepackage[T1]{fontenc}

\usepackage[utf8]{inputenc}

\usepackage{microtype}

\usepackage{inconsolata}
\usepackage{color,soul}
\usepackage{colortbl}
\usepackage{booktabs}
\usepackage{multirow}
\usepackage{listings}
\usepackage{arydshln}
\usepackage{graphicx}
\usepackage{amsmath}
\usepackage{tikz}
\usepackage{amsbsy}
\usepackage{amssymb}
\usepackage{tikz}
\usepackage{tcolorbox}\tcbuselibrary{skins}

\definecolor{mygreen}{HTML}{38761D}
\definecolor{myred}{HTML}{990000}
\definecolor{myblue}{HTML}{0073B7}
\definecolor{mydarkblue}{HTML}{2F5596}
\definecolor{myprefixgreen}{HTML}{009E73}
\definecolor{myorange}{HTML}{D55E00}
\definecolor{mygray}{HTML}{999999}
\definecolor{myyellow}{HTML}{DAA520}
\definecolor{mydarkgoldenrod}{HTML}{B8860B}
\definecolor{mypurple}{HTML}{6F42C1}
\definecolor{myolivegreen}{HTML}{556B2F}
\definecolor{mydarkslateblue}{HTML}{483D8B}
\definecolor{lightcoral}{HTML}{F08080}
\definecolor{lightslategray}{HTML}{778899}
\definecolor{cellcolor}{HTML}{DAA520}
\definecolor{tum-blue-brand}{RGB}{48, 112, 179}
\definecolor{theme_color}{RGB}{21, 87, 166}

\definecolor{bg_red}{HTML}{F9CECD}
\definecolor{up_red}{HTML}{B23333}
\definecolor{bg_green}{HTML}{DBE7C9}
\definecolor{up_green}{HTML}{789461}
\definecolor{bg_blue}{HTML}{d8e8fa}
\definecolor{up_blue}{HTML}{678bba}

\newcommand{\improve}[1]{{\color[HTML]{CB4335} $\uparrow$}}
\newcommand{\drop}[1]{{\color[HTML]{2E86C1} $\downarrow$}}
\newcommand{\best}[1]{{\color[HTML]{CB4335} $\checkmark$}}

\newcommand\RotText[1]{\rotatebox{90}{\parbox{1cm}{\centering#1}}}

\newcommand{\Adult}[0]{\textbf{\textcolor{myred}{\texttt{Adult}}}}
\newcommand{\GermanCredit}[0]{\textbf{\textcolor{myolivegreen}{\texttt{German Credit}}}}
\newcommand{\COMPAS}[0]{\textbf{\textcolor{mydarkblue}{\texttt{COMPAS}}}}

\newcommand{\ACC}[0]{\colorbox{lightcoral!15}{\texttt{ACC}}}
\newcommand{\F}[0]{\colorbox{bg_green!50}{\texttt{F1}}}
\newcommand{\SP}[0]{\colorbox{bg_blue!50}{\texttt{SP}}}
\newcommand{\EoO}[0]{\colorbox{myyellow!10}{\texttt{EoO}}}

\newcommand{\hlogo}{\raisebox{3.4pt}{\includegraphics[scale=0.0085]{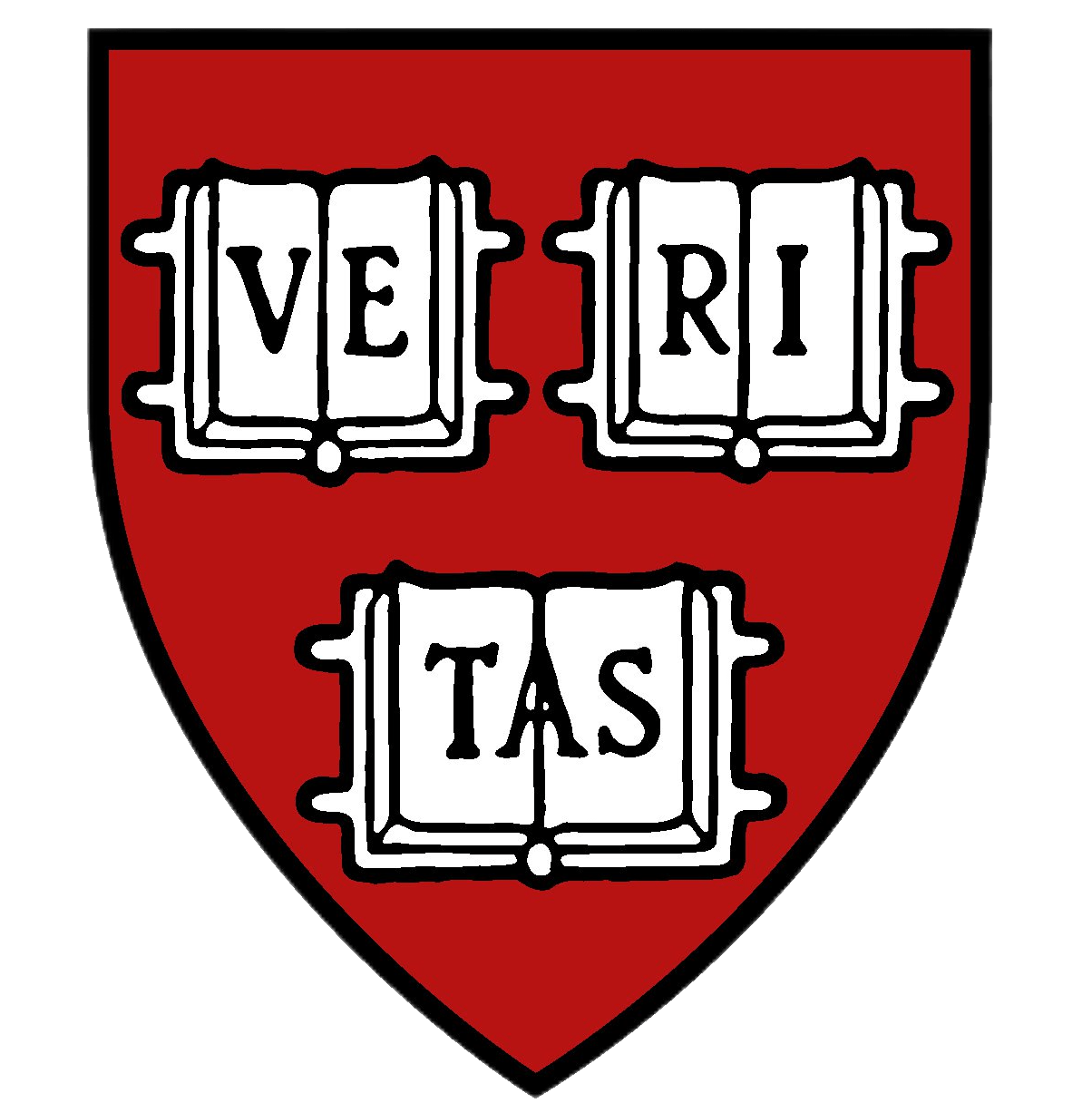}}}
\newcommand{\uitlogo}{\raisebox{2.6pt}{\includegraphics[scale=0.061]{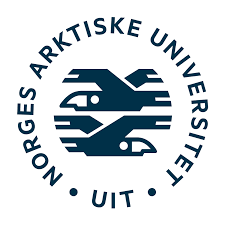}}}
\newcommand{\uiuclogo}{\raisebox{3.8pt}{\includegraphics[scale=0.125]{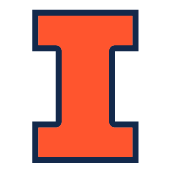}}}

\tcbset{prompt/.style={
    enhanced,
    size=fbox,
    boxrule=2.5pt,
    arc=2mm,
    auto outer arc,
    left=1pt,
    right=1pt,
    top=1pt,
    bottom=1pt,
    colframe=theme_color!90,
    colback=white,
}}

\title{Confronting LLMs with Traditional ML: Rethinking the Fairness of Large Language Models in Tabular Classifications}

\author{Yanchen Liu\hlogo\hspace{2.5pt} Srishti Gautam\uitlogo\hspace{2.5pt} Jiaqi Ma\uiuclogo\hspace{2.5pt} Himabindu Lakkaraju\hlogo\\
\hlogo Harvard University \hspace{1pt} \uitlogo UiT The Arctic University of Norway \\\uiuclogo University of Illinois Urbana-Champaign
\\
\texttt{yanchenliu@g.harvard.edu}\texttt{, srishti.gautam@uit.no,} \\\texttt{jiaqima@illinois.edu, }\texttt{hlakkaraju@hbs.edu}
}

\begin{document}
\maketitle

\begin{abstract}
Recent literature has suggested the potential of using large language models (LLMs) to make classifications for tabular tasks. However, LLMs have been shown to exhibit harmful social biases that reflect the stereotypes and inequalities present in society. To this end, as well as the widespread use of tabular data in many high-stake applications, it is important to explore the following questions: what sources of information do LLMs draw upon when making classifications for tabular tasks; whether and to what extent are LLM classifications for tabular data influenced by social biases and stereotypes; and what are the consequential implications for fairness?

Through a series of experiments, we delve into these questions and show that LLMs tend to inherit social biases from their training data which significantly impact their fairness in tabular classification tasks. Furthermore, our investigations show that in the context of bias mitigation, though in-context learning and finetuning have a moderate effect, the fairness metric gap between different subgroups is still larger than that in traditional machine learning models, such as Random Forest and shallow Neural Networks. This observation emphasizes that the social biases are inherent within the LLMs themselves and inherited from their pretraining corpus, not only from the downstream task datasets. Besides, we demonstrate that label-flipping of in-context examples can significantly reduce biases, further highlighting the presence of inherent bias within LLMs.
 
\end{abstract}

\section{Introduction}
Many recent works propose to use large language models (LLMs) for tabular tasks \citep{tabletSlack23, hegselmann2023tabllm}, where the tabular data is serialized as natural language and provided to LLMs with a short description of the task to solicit predictions. Despite the comprehensive examination of fairness considerations within conventional machine learning approaches applied to tabular tasks~\citep{aif360-oct-2018}, the exploration of fairness-related issues in the context of employing LLMs for tabular classifications remains a relatively underexplored domain.

Previous research has shown that LLMs, such as GPT-3 \citep{gpt3}, GPT-3.5, GPT-4~\citep{openai2023gpt4} can exhibit harmful social biases~\citep{Abid-Muslims-2021, Basta_Costa-jussà_Casas_2019}, 
which may even worsen as the models become larger in size \citep{askell2021general, Ganguli_2022}. 
These biases are a result of the models being trained on text generated by humans that presumably includes many examples of humans exhibiting harmful stereotypes and discrimination and reflects the biases and inequalities present in society~\citep{NIPS2016_Computer, zhao-etal-2017-men}, 
which can lead to the perpetuation of discrimination and stereotype~\citep{Abid-Muslims-2021, Bender_Gebru_McMillan-Major_Shmitchell_2021}.

Considering that tabular data finds extensive use in high-stakes domains \citep{grinsztajn2022why} where information is typically structured in tabular formats as a natural byproduct of relational databases~\citep{tabular_tnnls}, it is crucial to thoroughly examine the fairness implications of utilizing LLMs for classifications on tabular data. In this paper, we conduct a series of investigations centered around this critical problem, with the goal of discerning the underlying information sources upon which LLMs rely when making tabular classifications. Through this exploration, our investigation aims to ascertain whether, and to what degree, LLMs are susceptible to being influenced by social biases and stereotypes in the context of tabular data classifications.

Through experiments using \texttt{GPT-3.5} to make classifications for tabular data in a zero-shot setting, we demonstrate that LLMs exhibit significant social biases (Section \ref{sec:zero-shot}). This evidence confirms that LLMs inherit social biases from their pretraining corpus and tend to rely on these biases when making classifications for tabular data.

Furthermore, we demonstrate that providing \texttt{GPT-3.5} with few-shot examples (in-context learning) or finetuning them on the entire training datasets both exhibit moderate effects on bias mitigation (Sections \ref{subsec:regular_icl} and \ref{subsec:reg_ft}). Nevertheless, the achieved fairness levels remain below what is typically attained with traditional machine learning methods, including Random Forests and shallow Neural Networks, once again underscoring the presence of inherent bias in LLMs.

Moreover, our investigation further reveals that flipping the labels of the in-context few-shot examples significantly narrows the gap in fairness metrics across different subgroups, but comes at the expected cost of a reduction in the prediction performance. This finding, in turn, further emphasizes and reaffirms the indication of inherent bias present in LLMs (Section \ref{subsec:label_flipped}). Additionally, we further show that while resampling the training data is a known and effective method for reducing biases in traditional machine learning methods like Random Forests and shallow Neural Networks, it proves to be less effective when applied to \texttt{GPT-3.5} (Section \ref{subsec:resampling}). 

These collective findings demonstrate the significant influence of social biases on \texttt{GPT-3.5}'s performance in tabular classifications. These biases significantly undermine fairness and pose potential risks for using LLMs on tabular data, especially considering that tabular data is extensively used in high-stakes domains, highlighting the need for more advanced and tailored strategies to address these biases effectively. Straightforward methods like in-context learning and data resampling may not be sufficient in this context.

\section{Related work}
\paragraph{Fairness and Social Biases in LLMs}
Fairness is highly desirable for ensuring the credibility and trustworthiness of algorithms. It has been demonstrated that unfair algorithms can reflect societal biases in their decision-making processes~\citep{Bender_Gebru_McMillan-Major_Shmitchell_2021, bommasani2021opportunities}, primarily stemming from the biases present in their training data~\citep{doi:10.1126/science.aal4230, zhao-etal-2017-men}. LLMs, pre-trained on vast natural language datasets, are particularly susceptible to inheriting these social biases and have been shown to exhibit biases related to gender~\citep{lucy-bamman-2021-gender}, religion~\citep{religion} and language variants~\citep{ziems-etal-2023-multi, liu2023dada}. These social biases can lead to the perpetuation of discrimination and stereotype~\citep{Abid-Muslims-2021, Bender_Gebru_McMillan-Major_Shmitchell_2021, weidinger2021ethical}. While recent works have made strides in addressing these issues, there still exists a significant gap in comprehensively assessing fairness in LLMs and its mitigation strategies for tabular data.

\paragraph{Tabular Tasks and LLM for Tabular Data}
Tabular data extensively exist in many domains \citep{shwartz-ziv2021tabular}. Previous works propose to utilize self-supervised deep techniques for tabular tasks \citep{yin-etal-2020-tabert, arik2021tabnet}, which, however, still underperform ensembles of gradient-boosted trees in the fully supervised setting \citep{grinsztajn2022why}. This disparity in performance can be attributed to the locality, sparsity, and mixed data types of tabular data. In recent times, LLMs have undergone intensive training using vast amounts of natural language data, which has enabled them to exhibit impressive performance across various downstream tasks~\citep{gpt3, openai2023gpt4}, even with little or no labeled task data. Therefore, recent approaches by~\cite{hegselmann2023tabllm, tabletSlack23} suggest serializing the tabular data as natural language, which is provided to LLM along with a short task description to generate predictions for tabular tasks.

However, tabular data plays a crucial role in numerous safety-critical and high-stakes domains~\citep{tabular_tnnls, grinsztajn2022why}, which makes fairness particularly crucial when employing LLMs for making predictions on tabular data, especially considering the inherent social biases present in LLMs. Despite the importance, this still remains largely unexplored. To the best of our knowledge, we regard our work as one of the most comprehensive investigations into the fairness issues arising when using LLMs for classification on tabular data.

\paragraph{In-Context Learning}
Significant improvements for various tasks have been achieved by providing in-context examples to LLMs \citep{gpt3, liu-etal-2022-makes, liu-etal-2023-semantic}. However, previous research by \cite{min-etal-2022-rethinking, Wei2023LargerLM, lyu-etal-2023-z} illustrate that the effective performance of in-context learning largely hinges on semantic priors rather than learning the input-label mapping \citep{akyurek2022learning, xie2022an, von2023transformers} and the labels of the in-context examples might not play a crucial role in in-context learning, with flipped or random labels sometimes having minimal impact on performance. Despite these findings, the predominant focus of existing investigation of in-context learning remains on conventional natural language processing tasks \citep{zhao2021calibrate, min-etal-2022-rethinking,  wei2023symbol, Wei2023LargerLM}, largely overlooking the domain of tabular data. Furthermore, the fairness of in-context learning and the impact of flipped labels on this fairness is yet to be thoroughly investigated.

\section{Experimental Setup}

In this section, we begin by presenting an overview of the experimental setup utilized throughout this work.

\subsection{Models}
\label{subsec:models}
In our work, we focus our experiments on \texttt{GPT-3.5} (engine \textit{GPT-3.5-turbo}) - an LLM released by OpenAI, trained with instruction tuning \citep{sanh2022multitask, wei2022finetuned} and reinforcement learning from human feedback (RLHF) \citep{ouyang2022training}, aligning LLMs with human preferences.

Furthermore, we also compare this most common LLM with conventional machine learning models, in order to gain insight into the propagation of biases found within LLMs. This propagation of biases is likely mirrored in traditional models as well, consequently, offering valuable additional perspectives on the social biases inherent in the training of LLMs. For this, we employ two widely used traditional models for tabular data, i.e., Random Forests (\texttt{RF}) and the shallow Neural Network (\texttt{NN}) of 3 layers. We provide additional implementation details for these traditional models in Appendix~\ref{appendix:rf_nn_hp}.

\subsection{Datasets and Protected Attributes}
\label{subsec:datasets}
To explore the fairness of LLMs in making classifications for tabular data, we utilize the following three widely used tabular datasets for assessing the fairness of traditional ML models: \textit{Adult Income} (\textbf{\Adult{}}) Dataset \citep{adult}, \textbf{\GermanCredit{}} Dataset \citep{uci_repo}, and \textit{Correctional Offender Management Profiling for Alternative Sanctions} (\textbf{\COMPAS{}}) Dataset \citep{larson2016analyzed}. In this section, we give a brief introduction to each dataset and discuss its associated protected attributes.

\paragraph{Adult}
The \textit{Adult Income} dataset (\Adult{}) is extracted from the 1994 U.S. Census Bureau database. The task is to predict whether a person earns more than \$50,000 per year based on their profile data (\textit{greater than 50K} or \textit{less than or equal to 50K}). The original Adult Income Dataset contains 14 features. Following previous work \citep{tabletSlack23}, we retain only 10 features: \textit{``workclass''}, \textit{``hours per week''}, \textit{``sex''}, \textit{``age''}, \textit{``occupation''}, \textit{``capital loss''''}, \textit{``education''}, \textit{``capital gain''}, \textit{``marital status''}, and \textit{``relationship''}. Our analysis on \Adult{} primarily focuses on \textit{``sex''} as the protected attribute, and \textit{female} is acknowledged as a disadvantaged group.

\paragraph{German Credit}
The \GermanCredit{} dataset is used to classify individuals based on their profile attributes as good or bad credit risks (\textit{good} or \textit{bad}). The raw dataset comprises 20 attributes. Consistent with previous work, we only retain the following features: \textit{``age''}, \textit{``sex''}, \textit{``job''}, \textit{``housing''}, \textit{``saving accounts''}, \textit{``checking account''}, \textit{``credit amount''}, \textit{``duration''}, and \textit{``purpose''}. Same with \Adult{}, \textit{``sex''} is considered as a protected attribute in the \GermanCredit{} dataset and \textit{female} as the marginalized group.

\paragraph{COMPAS}
The \COMPAS{} dataset comprises the outcomes from the \textit{Correctional Offender Management Profiling for Alternative Sanctions} commercial algorithm, utilized to evaluate a convicted criminal's probability of reoffending. Known for its widespread use by judges and parole officers, \COMPAS{} has gained notoriety for its bias against African-Americans. The raw COMPAS Recidivism dataset contains more than 50 attributes. Following the approach of \citet{larson2016analyzed}, we perform necessary preprocessing, group \textit{``race''} into \textit{African-American} and \textit{Not African-American}, and only consider the features \textit{``sex''}, \textit{``race''}, \textit{``age''}, \textit{``charge degree''}, \textit{``priors count''}, \textit{``risk''} and \textit{``two year recid''} (target). We frame the task as predicting whether an individual will recidivate in two years (\textit{Did Not Reoffend} or \textit{Reoffended}) based on their demographic and criminal history. For the \COMPAS{} dataset, we consider \textit{``race''} as the protected attribute.

A detailed description for each feature of the considered datasets is provided in Appendix \ref{appendix:features}.

\subsection{Serialization and Prompt Templates}
\label{subsec:prompt}
To employ the LLM for making classifications on these tabular datasets, each data point is first serialized as text. Following previous works on LLM for tabular classifications \citep{hegselmann2023tabllm, tabletSlack23}, we format the feature names and values into strings as ``$f_1: x_1, \ldots, f_d: x_d$'', and prompt to LLM along with a task description, as illustrated following:
\vspace{5pt}

\begin{tcolorbox}[prompt, title = {{Prompt 1. Prompt Template for \texttt{Adult} Dataset.}}]
{
\vspace{-0.2cm}
\begin{lstlisting}[
    breaklines=true,
    postbreak=\mbox{\hspace{-0.7cm}},
    basicstyle=\color{myblue}\ttfamily\footnotesize,
    columns=fullflexible,
    escapeinside={(*}{*)}
]
You must predict if income exceeds $50K/yr. Answer with one of the following: greater than 50K | less than or equal to 50K.
\end{lstlisting}
}
{
\vspace{-0.4cm}
\begin{lstlisting}[
    breaklines=true,
    postbreak=\mbox{\hspace{-0.7cm}},
    basicstyle=\color{mygreen}\ttfamily\footnotesize,
    columns=fullflexible,
    escapeinside={(*}{*)}
]
Example 1 -
workclass: Private
hours per week: 20
sex: Male
age: 17
occupation: Other-service
capital loss: 0
education: 10th
capital gain: 0
marital status: Never-married
relationship: Own-child
Answer: less than or equal to 50K
...
\end{lstlisting}
}
{
\vspace{-0.3cm}
\begin{lstlisting}[
    breaklines=true,
    postbreak=\mbox{\hspace{-0.7cm}},
    basicstyle=\color{myred}\ttfamily\footnotesize,
    columns=fullflexible,
    escapeinside={(*}{*)}
]
workclass: Private
hours per week: 40
sex: Female
age: 24
occupation: Sales
capital loss: 0
education: Some-college
capital gain: 0
marital status: Never-married
relationship: Own-child
Answer:
\end{lstlisting}
\vspace{-0.25cm}
}\end{tcolorbox}
\vspace{5pt}

The example above is from the \Adult{} dataset, where text in \textcolor{myblue}{blue} denotes the task description, text in \textcolor{mygreen}{green} indicates optional few-shot examples (only used in in-context learning setting), and text in \textcolor{myred}{red} is the test example. We provide the prompt templates for the other two datasets in Appendix \ref{appendix:prompts}.

\subsection{Evaluation Metrics}
\label{subsec:metrics}
To assess fairness in the aforementioned datasets, we examine the disparity between different subgroups of protected attributes using the following common fairness metrics: accuracy (\ACC{}), F1 score (\F{}), statistical parity (\SP{}), and equality of opportunity (\EoO{}). Here, we briefly explain each evaluation metric.

\paragraph{Accuracy and F1} As the most basic metric, assessing accuracy among different subgroups ensures that the model delivers consistent performance across all groups, without undue favor to any particular subgroups.

Considering that the evaluated datasets may be imbalanced, especially among different subgroups, the \F{} Score computes the harmonic mean of precision and recall, offering a balanced perspective between these two metrics.

\paragraph{Statistical Parity} Statistical parity is attained when \textit{positive} decision outcomes (e.g., being predicted as good credit risk) are independent of the protected attributes. This metric assesses whether different subgroups receive similar treatment from the model. For each subgroup $\boldsymbol z_i$ of each protected attribute $Z$, we calculate its statistical parity as:

{
\fontsize{10pt}{12pt}\selectfont
\begin{equation*}
    P(\hat{Y} = 1|Z = \boldsymbol z_i).
\end{equation*}
}

Then we calculate the difference of Statistical Parity ($d_{\SP{}}$) of this protected attribute as:
\vspace{-0.4cm}

{
\fontsize{10pt}{12pt}\selectfont
\begin{equation*}
    d_{\SP{}} = \overset{\text{\textcolor{up_red!90}{\textbf{\fontsize{8pt}{9pt}\selectfont{statistical parity of $\boldsymbol z_1$}}}}}{\colorbox{bg_red}{$P(\hat{Y} = 1 \mid Z = \boldsymbol z_1)$}} - \overset{\text{\textcolor{up_green}{\textbf{\fontsize{8pt}{9pt}\selectfont{statistical parity of $\boldsymbol z_2$}}}}}{\colorbox{bg_green}{$P(\hat{Y} = 1 \mid Z = \boldsymbol z_2)$}},
\end{equation*}
}

where \textcolor{up_red!90}{$\boldsymbol z_1$} is the \textbf{\textcolor{up_red!90}{minority group}} and \textcolor{up_green}{$\boldsymbol z_2$} is the \textbf{\textcolor{up_green}{majority}}.

\paragraph{Equality of Opportunity} Equality of opportunity requires that qualified individuals have an equal chance of being correctly classified by the model, regardless of their membership in a protected group. This metric ensures equal \textit{true positive} rates between different subgroups, providing equal opportunities for each subgroup. Similar to statistical parity, for equality of opportunity, we calculate the difference of Equality of Opportunity ($d_{\EoO{}}$) as:
\vspace{-0.4cm}

{
\fontsize{10pt}{12pt}\selectfont
\begin{align}
    d_{\EoO{}} &= \overset{\text{\textcolor{up_red!90}{\textbf{\fontsize{8pt}{9pt}\selectfont{equality of opportunity of $\boldsymbol z_1$}}}}}{\colorbox{bg_red}{$P(\hat{Y} = 1 \mid Y = 1, Z = \boldsymbol z_1)$}} \nonumber \\
    &\quad - \overset{\text{\textcolor{up_green}{\textbf{\fontsize{8pt}{9pt}\selectfont{equality of opportunity of $\boldsymbol z_2$}}}}}{\colorbox{bg_green}{$P(\hat{Y} = 1 \mid Y = 1, Z = \boldsymbol z_2)$}}. \nonumber
\end{align}
}

Each of these metrics offers a different perspective on fairness. For each subgroup from each protected attribute, we will compute every aforementioned metric. A model demonstrating good fairness should show minimal gaps in these fairness metrics between different subgroups. Considering them together can provide a more comprehensive evaluation of the model's fairness across different subgroups, ensuring that individuals are not unfairly disadvantaged based on their membership in a protected group.

\section{Zero-Shot Prompting for Tabular Data}

\begin{figure*}[!ht]
  \centering
  \includegraphics[width=\textwidth]{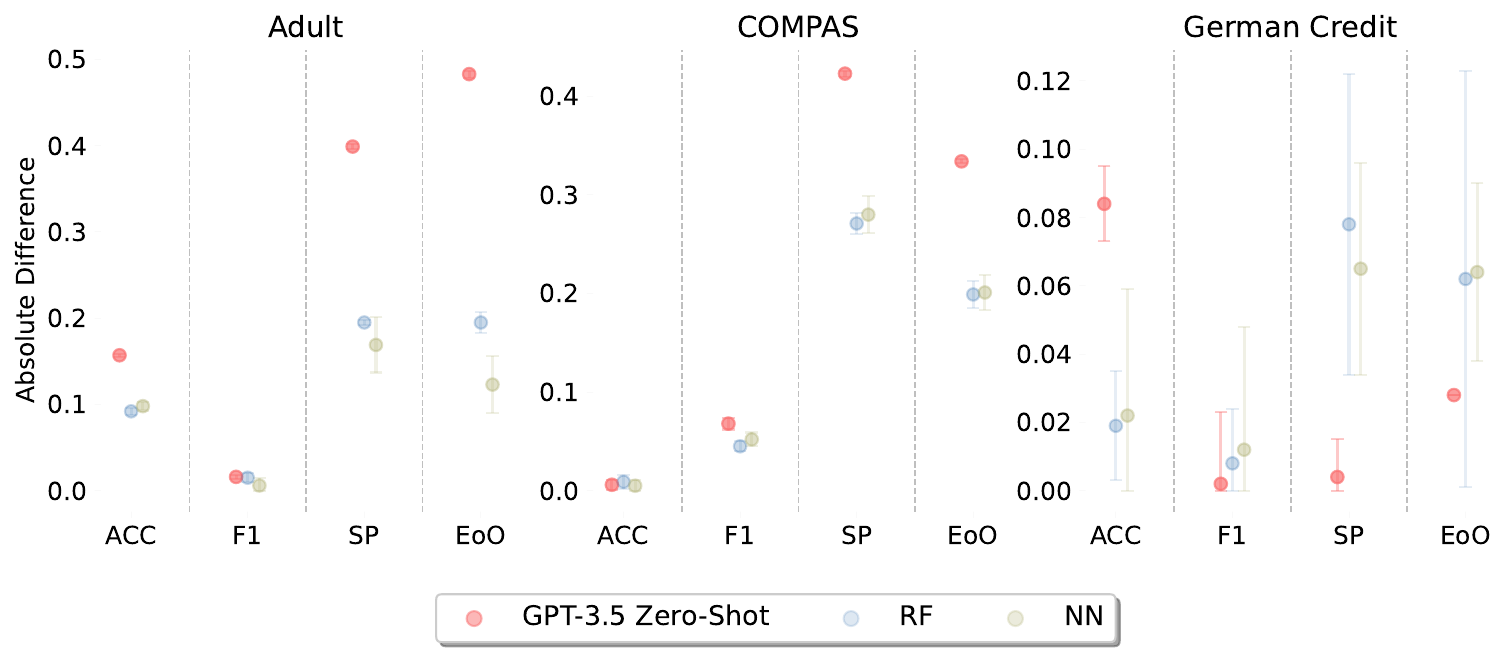}
  \caption{\textbf{Fairness Comparison of \texttt{GPT-3.5 Zero-Shot} Prompting with \texttt{RF} and \texttt{NN}.} We present the absolute differences in \ACC{}, \F{}, \SP{}, and \EoO{} between subgroups of protected attributes across three datasets: \textit{``sex''} for \Adult{} and \GermanCredit{}, and \textit{``race''} for \COMPAS{}. Notable fairness gaps between protected and non-protected subgroups are observed for \texttt{GPT-3.5} in a zero-shot prompting manner (\texttt{GPT-3.5 Zero-Shot}), which are significantly larger compared to those observed with \texttt{RF} and \texttt{NN}, except for the \GermanCredit{} dataset.
}
  \label{fig:zero-shot}
  \vspace{5pt}
\end{figure*}

\begin{table*}[htbp]
\centering
\resizebox{0.7 \textwidth}{!}{
\begin{tabular}{l|cccc}
& $d_{\ACC{}}$ & $d_{\F{}}$ & $d_{\SP{}}$ & $d_{\EoO{}}$ \\
\hline\hline
\multicolumn{5}{c}{\Adult{}} \\
\cdashline{1-5}
w/ protected attr. & 0.157\textsubscript{0.002} & -0.016\textsubscript{0.002} & -0.399\textsubscript{0.003} & -0.483\textsubscript{0.004} \\
w/o protected attr. & 0.106\textsubscript{0.002} & -0.021\textsubscript{0.001} & -0.287\textsubscript{0.002} & -0.273\textsubscript{0.002} \\
\cdashline{1-5}
\multicolumn{5}{c}{\COMPAS{}} \\
\cdashline{1-5}
w/ protected attr. & -0.006\textsubscript{0.005} & 0.068\textsubscript{0.006} & -0.423\textsubscript{0.003} & -0.334\textsubscript{0.002} \\
w/o protected attr. & -0.002\textsubscript{0.003} & 0.058\textsubscript{0.004} & -0.239\textsubscript{0.002} & -0.157\textsubscript{0.004} \\
\hline
\end{tabular}
}
\caption{\textbf{Fairness Comparison of \texttt{GPT-3.5 Zero-Shot} Prompting with and without Protected Attributes.} When protected attributes are removed from the input, there is a notable decrease in fairness gaps, especially in \SP{} and \EoO{}. This observation provides further evidence of the inherent bias present in \texttt{GPT-3.5}.}
\label{tab:zeroshot}
\end{table*}

\label{sec:zero-shot}
To explore the fairness of LLMs when making classifications on 
tabular data, we first conduct experiments in a zero-shot setting. We assess the fairness metrics of the outcomes and examine whether \texttt{GPT-3.5} without any finetuning or few-shot examples would be influenced by social biases and stereotypes for tabular classification. We run all the experiments 5 times and compute the mean and standard deviation.

\subsection{Regular Zero-Shot Prompting}

In Figure \ref{fig:zero-shot}, we compare the absolute differences in four evaluated fairness metrics between subgroups of protected attributes for \texttt{GPT-3.5} in a zero-shot manner (\texttt{GPT-3.5 Zero-Shot}) with \texttt{RF} and \texttt{NN} on the \Adult{}, \GermanCredit{}, and \COMPAS{} datasets, respectively. The evaluted metrics include accuracy (\ACC{}), F1 score (\F{}), statistical parity (\SP{}), and equality of opportunity (\EoO{}). For the \Adult{} and \GermanCredit{} datasets, the subgroups \textit{female} and \textit{male} are assessed regarding the protected attribute \textit{``sex''}, identifying \textit{female} as a disadvantaged group. In the \COMPAS{} dataset, we evaluate \textit{``race''} as protected attributes, recognizing \textit{African American (AA)} as the disadvantaged group.
Due to space limitations, we included the concrete numbers in Tables \ref{tab:adult} to \ref{tab:compas} in Appendix \ref{appendix:full}.

It is notable that when directly utilizing LLMs to make classifications for tabular data, without any fine-tuning or in-context learning, a significant fairness metric gap between the protected and non-protected groups is observed for \texttt{GPT-3.5} (highlighted in \textcolor{myred}{red} in Appendix \ref{appendix:full}). For example, the \EoO{} difference between \textit{male} and \textit{female} on the \textit{\Adult{}} dataset reaches 0.483, indicating a substantial disadvantage for the \textit{female} group. Additionally, when compared with conventional models like \texttt{RF} and \texttt{NN}, the biases in zero-shot predictions made by \texttt{GPT-3.5} are significantly larger when applied to the \Adult{} and \COMPAS{} dataset. This observation suggests the presence of inherent gender and race biases within \texttt{GPT-3.5}.

Exceptionally, \texttt{GPT-3.5} is extremely biased for \GermanCredit{} dataset where it classifies almost everything into \textit{good credit} class in the zero-shot setting, thus rendering the difference in \SP{} and \EoO{} for both subgroups to be near 0. The accuracy for each subgroup is near 50\%, performing similar to random guessing. The possible reason might be that the \GermanCredit{} dataset is too challenging for making tabular classifications with LLMs (especially, since the features of \GermanCredit{} are ambiguous and vague). This also suggests that, when using \texttt{GPT-3.5} to make predictions on tabular data, a potential description of table feature names is favorable.

\subsection{Zero-Shot Prompting with Protected Attributes Removed}
To further demonstrate inherent social biases in LLMs, we compare fairness gaps in zero-shot classifications using \texttt{GPT-3.5} under two conditions: one including protected attributes and the other with these attributes removed. As shown in Table \ref{tab:zeroshot}, fairness gaps notably decrease, particularly in \SP{} and \EoO{}, when protected attributes are excluded from the input. This observation further confirms the presence of inherent bias in \texttt{GPT-3.5}.

Taken together, these results demonstrate the tendency of \texttt{GPT-3.5} to rely on social biases and stereotypes inherited from their pretraining corpus when applied to tabular data. This suggests that using LLMs for classifications on tabular data may incur significant fairness risks, potentially disproportionately disadvantaging marginalized communities and perpetuating societal biases and stereotypes. Given the widespread use of tabular data in high-stakes contexts, these findings raise serious concerns about the potential for harm.

\begin{figure*}[!ht]
  \centering
  \includegraphics[width=\textwidth]{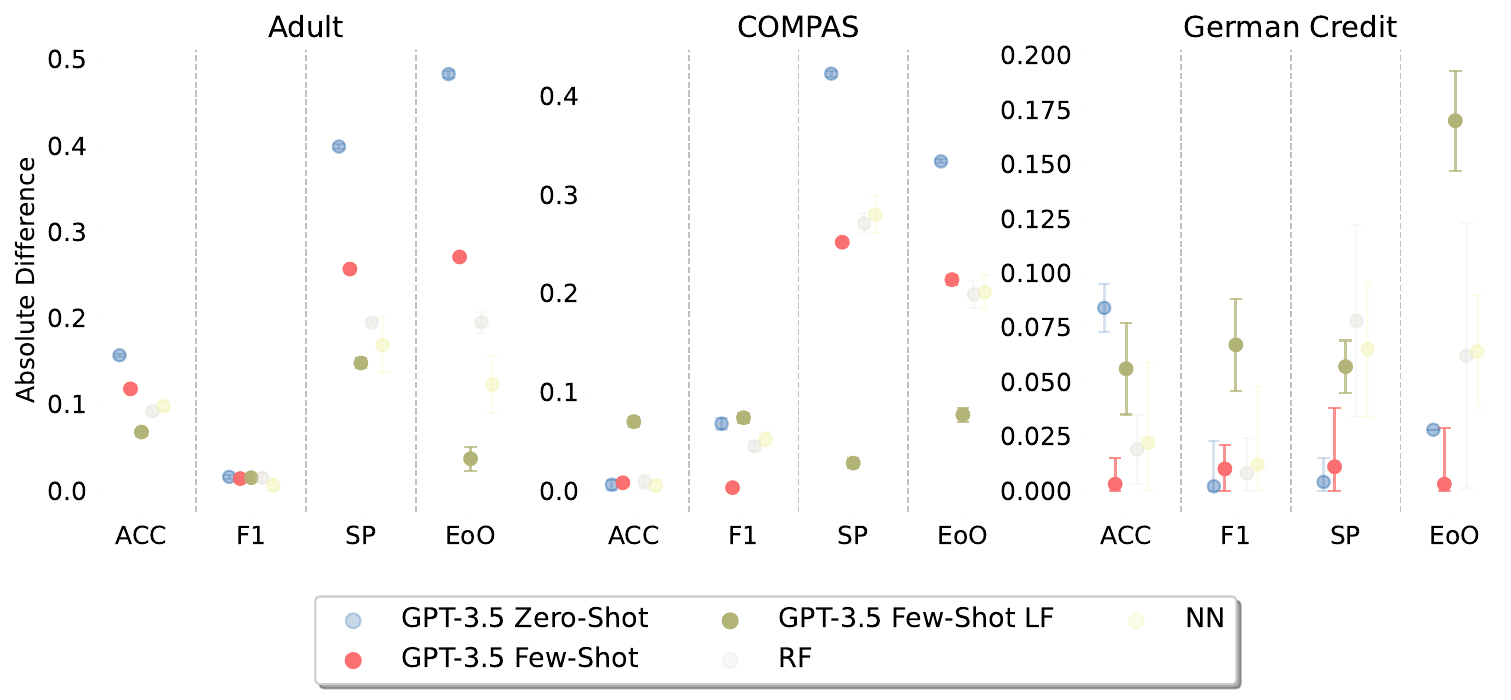}
  \caption{\textbf{Fairness Comparison of \texttt{GPT-3.5 Few-Shot} Prompting.} We compare the absolute differences in \ACC{}, \F{}, \SP{}, and \EoO{} between subgroups of protected attributes across the \Adult{}, \COMPAS{} and and \GermanCredit{} datasets for GPT-3.5 in three settings: zero-shot prompting (\texttt{GPT-3.5 Zero-Shot}), few-shot prompting (\texttt{GPT-3.5 Few-Shot}), and few-shot prompting with flipped labels (\texttt{GPT-3.5 Few-Shot LF}). Incorporating few-shot examples (\texttt{GPT-3.5 Few-Shot}) can partially reduce the inherent biases in \texttt{GPT-3.5}, but it cannot completely eliminate them. The fairness gap persists and is greater than that observed in \texttt{RF} and \texttt{NN}. Furthermore, label-flipped few-shot examples (\texttt{GPT-3.5 Few-Shot LF}) can effectively reverse the bias effects, further narrowing the fairness gaps.
}
  \label{fig:few-shot}
\end{figure*}
\section{Few-Shot Prompting for Tabular Data}

\label{sec:few-shot}
As demonstrated in Section \ref{sec:zero-shot}, employing \texttt{GPT-3.5} for classifications on tabular data reveals significant social biases in a zero-shot setting. Instead of directly utilizing \texttt{GPT-3.5} for zero-shot tabular classifications, this section explores whether including few-shot examples during prompting will reduce or amplify these biases. To delve deeper into the influence of few-shot examples during in-context learning (ICL), we not only consider the regular ICL approach as detailed in Section \ref{subsec:regular_icl}, but we also experiment by flipping the labels of the few-shot examples to further examine their effect on the biases, as discussed in Section \ref{subsec:label_flipped}.

Again, for robustness, each experiment is conducted 5 times, with the mean and standard deviation reported. We present the fairness gap in Figure \ref{fig:few-shot} and the complete results in Appendix \ref{appendix:full}.

\subsection{Regular In-Context Learning}
\label{subsec:regular_icl}
Previous works have demonstrated that LLMs can learn the input-label mappings in context \citep{akyurek2022learning, xie2022an, von2023transformers}. However, the influence of in-context learning on fairness has not been thoroughly examined. For in-context learning, the test example and task description, along with a few-shot examples, are provided to the LLMs for generating the final predictions. The few-shot examples are inserted before the test example in the prompt, as outlined in Section \ref{subsec:prompt}. We set the number of in-context examples as 50. For each dataset, we randomly select the in-context examples from the training set for each test example.

In Tables \ref{tab:adult}-\ref{tab:compas}, we demonstrate that for two of the evaluated datasets (except for \COMPAS{}), the incorporation of few-shot examples brings about performance improvements.
Additionally, in Figure \ref{fig:few-shot}, we observe that incorporating few-shot examples into prompting reduces the fairness metric gap between different subgroups. However, a significant fairness issue still persists. Moreover, for the \Adult{} and \COMPAS{} datasets, the disparity in fairness metrics of in-context learning is more notable when compared to traditional models \texttt{RF} and \texttt{NN}.
This highlights the inherent biases embedded within LLMs, which are not solely derived from the task datasets.

\subsection{Label Flipping}
\label{subsec:label_flipped}
To delve deeper into the sources of biases within \texttt{GPT-3.5}, we further examine the impact of the labels of in-context examples on fairness. As demonstrated in Figure \ref{fig:few-shot}, label flipping significantly reduces biases across all evaluated datasets.
For all datasets, the difference in statistical parity (\SP{}) and equality of opportunity (\EoO{}) is minimized with label-flipped ICL.
For example, the absolute gap of \EoO{} on the \Adult{} dataset decreases from 0.483 in zero-shot prompting to 0.037, almost completely eliminating the bias.
These findings further corroborate the existence of inherent biases in \texttt{GPT-3.5}. 

However, flipped labels lead to a significant drop in classification performance.  Though previous research suggests that the effectiveness of ICL predominantly stems from semantic priors, rather than learning the input-label mappings \citep{min-etal-2022-rethinking, Wei2023LargerLM} and demonstrates that the performance of ICL is barely affected even with flipped or random labels for in-context examples, the focus of these works lies mainly on traditional natural language processing tasks. In contrast, we observe that the labels of in-context examples hold substantial influence over the performance in our unique setup, where \texttt{GPT-3.5} are deployed for classifications on tabular data. This could be attributed to the limited exposure of these models to tabular data during pre-training, thereby amplifying the role of input-label mapping of in-context examples.

\begin{figure*}[!ht]
  \centering
  \includegraphics[width=\textwidth]{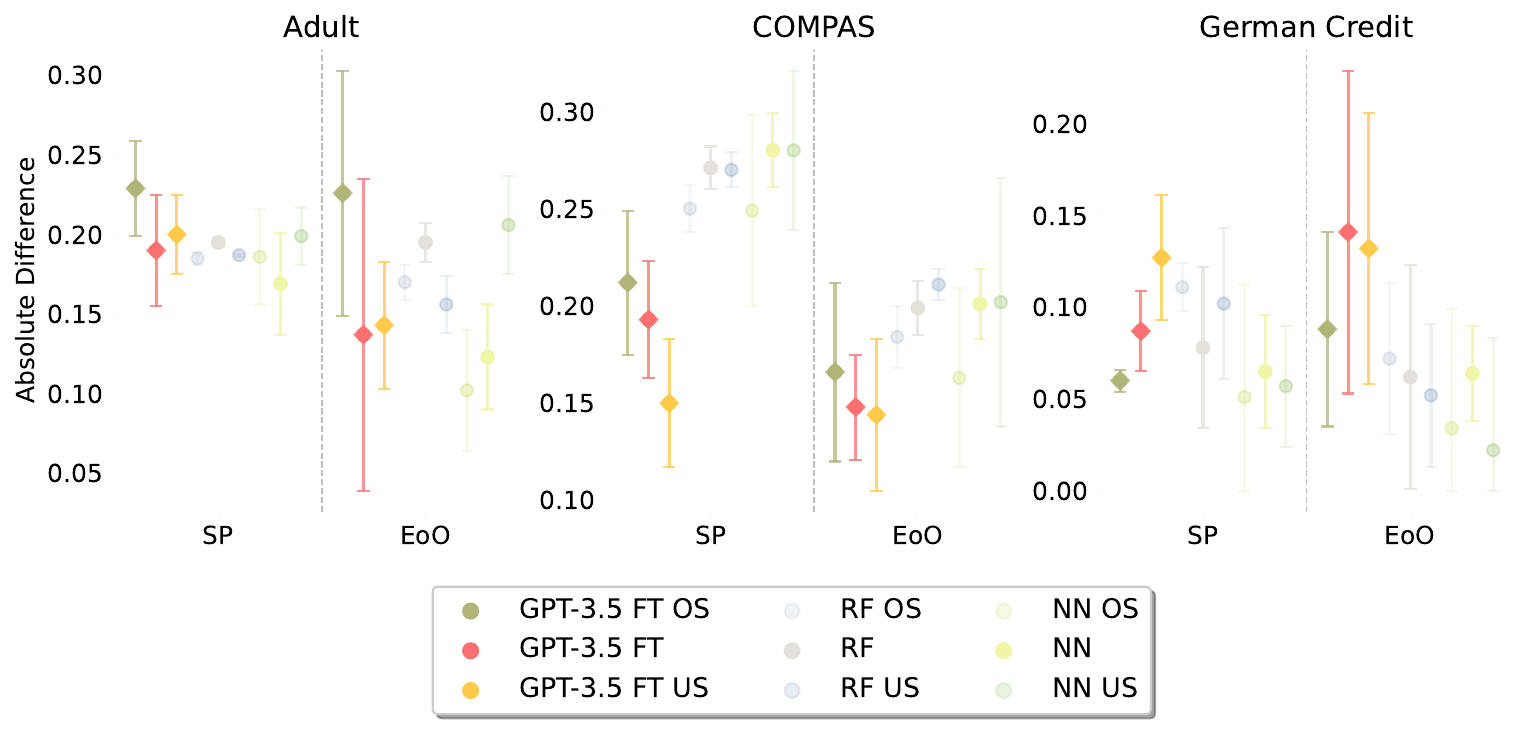}
  \caption{\textbf{Fairness Comparison of \texttt{GPT-3.5 Finetuning} with \texttt{RF} and \texttt{NN} Finetuning.} We compare the absolute differences in fairness metrics - \ACC{}, \F{}, \SP{}, and \EoO{} - between subgroups of protected attributes across the \Adult{}, \COMPAS{} and \GermanCredit{} datasets for the finetuned \texttt{GPT-3.5} models using three different approaches: finetuning on the entire training data (\texttt{GPT-3.5 FT}), oversampled data (\texttt{GPT-3.5 FT OS}), and undersampled data (\texttt{GPT-3.5 FT US}). While finetuning \texttt{GPT-3.5} on task datasets (\texttt{GPT-3.5 FT}) can mitigate inherent social biases to some extent, the application of data resampling techniques (\texttt{GPT-3.5 FT OS} and \texttt{GPT-3.5 FT US}) during the finetuning process does not consistently yield similar results for LLMs compared to the typical mitigation observed in traditional machine learning models.
  }
  \label{fig:finetuning}
\end{figure*}
\section{Finetuning for Tabular Data}

\label{sec:finetuning}
\subsection{Regular Finetuning}
\label{subsec:reg_ft}
Finally, we extend our investigation to assess if finetuning the models on the entire training set could aid in diminishing the social biases in LLMs. 
For \texttt{GPT-3.5}, fine-tuning is executed using the publicly released API from OpenAI. For \texttt{RF} and \texttt{NN}, we provide the training details in Appendix \ref{appendix:rf_nn_hp}. We still run all the experiments 5 times and compute the mean and standard deviation. 
In Figure \ref{fig:finetuning}, we show that finetuning (\texttt{GPT-3.5 FT}) effectively reduces unfairness in all datasets, making them comparable and sometimes significantly better in terms of \SP{} and \EoO{} when compared to \texttt{RF} and \texttt{NN}. For example, the absolute difference in \EoO{} after finetuning on the \Adult{} dataset is 0.0714, which is lower than the 0.123 difference of an \texttt{NN}.

\subsection{Resampled Finetuning}
\label{subsec:resampling}
Resampling is a method frequently utilized to enhance fairness in traditional machine learning model training, particularly in scenarios where there is a significant class imbalance or bias in the data.
We further explore the potential of resampling the task datasets to reduce the fairness gap for LLMs.
Specifically, we evaluate two approaches: oversampling the minority group (\texttt{OS}) and undersampling the majority group (\texttt{US}). As depicted in Figure \ref{fig:finetuning}, resampling fails to mitigate the social biases in \texttt{GPT-3.5} when making tabular classification, even though we demonstrate that oversampling generally reduces social biases for both \texttt{RF} and \texttt{NN}, except for a few instances such as oversampling in \texttt{NN} for \Adult{} dataset worsens the fairness.

Our finetuning experiments show that the social biases inherited from \texttt{GPT-3.5}'s pre-training corpus, which are noticeably evident when making classifications on tabular data, can sometimes be mitigated through finetuning. Nevertheless, unlike the consistent outcomes typically seen in traditional machine learning models, data resampling does not consistently produce similar results for finetuning LLMs.
\section{Conclusion}
In this work, we thoroughly investigate the under-explored problem of fairness of large language models (LLMs) for tabular tasks.  Our study unfolds in several phases.

Initially, we assess the inherent fairness displayed by GPT-3.5, comparing their performance in zero-shot learning scenarios against traditional machine learning models like random forests (RF) and shallow neural networks (NN). Furthermore, we investigate how GPT-3.5 learns and propagates social biases when subjected to few-shot in-context learning, label-flipped in-context learning, finetuning, and data resampling techniques.

Our discoveries shed light on several key insights. We find that GPT-3.5 tends to heavily rely on the social biases inherited from their pre-training data when making predictions, which is a concerning issue. Moreover, we observe that few-shot in-context learning can partially mitigate the inherent biases in GPT-3.5, yet it cannot entirely eliminate them. A significant fairness metric gap between different subgroups persists and exceeds that observed in RF and NN. This observation underscores the existence of biases within the LLMs themselves, beyond just the task datasets. Additionally, label-flipping applied to the few-shot examples effectively reverses the effects of bias, again corroborating the existence of inherent biases in GPT-3.5. However, as expected, this leads to a loss in the classification performance.
Besides, our work reveals that while fine-tuning can sometimes improve the fairness of GPT-3.5, data resampling does not consistently yield the same results, unlike what is typically observed in traditional machine learning models. This underscores the need for the development of more effective strategies to mitigate the bias inherent in LLMs and ensure their fairness when deployed in real-world applications.

\section*{Limitations}
It is important to note that our study exclusively focuses on the GPT-3.5 model. Consequently, our conclusions are representative of GPT-3.5 alone and cannot be extrapolated to other LLMs, which might exhibit different behaviors or biases. This focus on a single model thus restricts the broader applicability of our findings.

Furthermore, for each experiment, we employed only one type of prompt. This approach limits the generalizability of our conclusions, as different prompts might yield varying results. The use of a singular prompt type does not capture the full spectrum of possible interactions and outcomes that might be observed with a diverse range of prompting strategies.

Looking ahead, we plan to broaden the scope of our research. This expansion will include experimenting with additional models beyond GPT-3.5, thus offering a more comprehensive understanding of fairness for different LLMs. We also intend to explore a variety of prompting strategies, such as Chain of Thought (CoT) prompting, to assess how different methods may impact model bias and fairness. These future endeavors aim to provide a more nuanced and thorough exploration of the capabilities and limitations of LLMs in the context of fairness.

\bibliography{anthology,custom}
\bibliographystyle{acl_natbib}

\appendix
\clearpage

\appendix
\section{Dataset Description}
\label{appendix:features}
We provide a detailed description of each feature from the datasets evaluated in our paper.

\subsection{Adult}

The original Adult Income Dataset contains 14 features and the target \textit{Income}, as described in Table \ref{tab:adult_features}. Following prior work \citep{tabletSlack23}, we omit \textit{Education-Num} and \textit{Fnlwgt} as they are not crucial for income prediction, along with \textit{Race} and \textit{Native-Country}, to center our attention on \textit{Sex} as the protected attribute.

\begin{table*}[!ht]
\centering
\small
\begin{tabular}{|c|p{4cm}|p{6cm}|}
\hline
\textbf{Feature} & \textbf{Type} & \textbf{Description} \\
\hline
\hline
Age & Continuous & Represents the age of an individual. \\
\hline
Workclass & Categorical & Indicates the type of employment, such as private, self-employed, or government. \\
\hline
\textit{Fnlwgt} & Continuous & Stands for ``final weight'' and is a numerical value used in sampling for survey data. \\
\hline
Education & Categorical & Specifies the highest level of education attained by the individual, such as high school, bachelor's degree, etc. \\
\hline
\textit{Education-Num} & Continuous & Represents the numerical equivalent of the education level. \\
\hline
Marital-Status & Categorical & Describes the marital status of the individual, including categories like married, divorced, or single. \\
\hline
Occupation & Categorical & Indicates the occupation of the individual, such as managerial, technical, or clerical work. \\
\hline
Relationship & Categorical & Specifies the individual's role in the family, such as husband, wife, or child. \\
\hline
Race & Categorical & Represents the individual's race or ethnic background. \\
\hline
Sex & Categorical & Indicates the gender of the individual, either male or female. \\
\hline
Capital-Gain & Continuous & Refers to the capital gains, which are profits from the sale of assets, of the individual. \\
\hline
Capital-Loss & Continuous & Represents the capital losses, which are losses from the sale of assets, of the individual. \\
\hline
Hours-Per-Week & Continuous & Denotes the number of hours worked per week by the individual. \\
\hline
\textit{Native-Country} & Categorical & Specifies the native country or place of origin of the individual. \\
\hline
Income (target) & Binary & The target variable indicating whether an individual's income exceeds a certain threshold, typically \$50,000 per year. \\
\hline
\end{tabular}
\caption{Features in the original \textbf{Adult} dataset. Those not used in our work are shown in \textit{italics}.}
\label{tab:adult_features}
\end{table*}

\subsection{German Credit}

The original German Credit Dataset contains 20 features, as detailed in Table \ref{tab:german-credit-features}. For simplicity and consistency with prior work, only the features not shown in \textit{italics} are retained in our work. Furthermore, we extract \textit{Sex} as an additional protected attribute from the \textit{Personal Status and Sex} feature.

\begin{table*}[!ht]
\centering
\small
\begin{tabular}{|c|p{3cm}|p{6cm}|}
\hline
\textbf{Feature} & \textbf{Type} & \textbf{Description} \\
\hline
\hline
Credit Amount & Continuous & The amount of credit requested by the applicant. \\
\hline
Duration & Continuous & The duration of the credit in months. \\
\hline
\textit{Installment Rate} & Ordinal & The installment rate in percentage of disposable income. \\
\hline
\textit{Residence Since} & Ordinal & The number of years the applicant has lived at their current residence. \\
\hline
Age & Continuous & The age of the applicant. \\
\hline
\textit{Number of Existing Credits} & Ordinal & The number of existing credits at this bank. \\
\hline
\textit{Number of Dependents} & Ordinal & The number of dependents of the applicant. \\
\hline
Checking Account Status & Categorical & The status of the applicant's checking account, such as ``no checking, ``\textless 0 DM,'' ``0-200 DM,'' or ``no known checking.'' \\
\hline
\textit{Credit History} & Categorical & The credit history of the applicant, including categories like ``critical/other existing credit,'' ``existing paid,'' ``delayed previously,'' etc. \\
\hline
Purpose & Categorical & The purpose of the credit, such as ``radio/tv,'' ``education,'' ``new car,'' etc. \\
\hline
Savings Account & Categorical & The status of the applicant's savings account/bonds, including categories like ``unknown/none,'' ``\textless 100 DM,'' ``500-1000 DM,'' etc. \\
\hline
\textit{Employment Since} & Categorical & The duration of the applicant's current employment, such as ``unemployed,'' ``\textless 1 year,'' ``4-7 years,'' etc. \\
\hline
\textit{Personal Status and Sex} & Categorical & The personal status and sex of the applicant, including categories like ``male single,'' ``female div/dep/mar,'' etc. \\
\hline
\textit{Other Debtors/Guarantors} & Categorical & Indicates the presence of other debtors/guarantors, such as ``none,'' ``guarantor,'' ``co applicant.'' \\
\hline
\textit{Property} & Categorical & Describes the type of property owned by the applicant, such as ``real estate,'' ``life insurance,'' ``car or other,'' etc. \\
\hline
\textit{Other Installment Plans} & Categorical & The presence of other installment plans. \\
\hline
Housing & Categorical & The housing situation of the applicant, such as ``own,'' ``for free,'' and ``rent.'' \\
\hline
Job & Categorical & The type of job held by the applicant, including categories like ``skilled,'' ``unskilled resident,'' ``high qualif/self emp/mgmt,'' etc. \\
\hline
\textit{Telephone} & Binary & Indicates whether the applicant has a telephone (yes/no). \\
\hline
\textit{Foreign Worker} & Binary & Indicates whether the applicant is a foreign worker (yes/no). \\
\hline
Risk (target) & Binary & The target variable indicating credit risk (good/bad). \\
\hline
\end{tabular}
\caption{Features in the original \textbf{German Credit} dataset. Those not used in our work are shown in \textit{italics}. Additionally, from the original feature \textit{Personal Status and Sex}, we extract \textit{Sex} as a protected attribute.}
\label{tab:german-credit-features}
\end{table*}

\subsection{COMPAS}
The raw COMPAS Recidivism dataset contains more than 50 attributes. Following the approach of \citet{larson2016analyzed}, we carry out the necessary preprocessing. More specifically, we group the \textit{race} attribute into \textit{African-American} and \textit{Not African-American}, and consider only the features \textit{sex}, \textit{race}, \textit{age}, \textit{charge degree}, \textit{priors count}, \textit{risk}, and \textit{two-year recid} (target). We frame the task as predicting whether an individual will recidivate within two years (\textit{Did Not Reoffend} or \textit{Reoffended}), based on their demographic and criminal history. Due to page limitations, we provide descriptions for only the features used in our work in Table \ref{tab:compas-features}.

\begin{table*}[!ht]
\centering
\small
\begin{tabular}{|c|p{4cm}|p{6cm}|}
\hline
\textbf{Feature} & \textbf{Type} & \textbf{Description} \\
\hline
\hline
Sex & Categorical & The gender of the individual. \\
\hline
Race & Categorical & The race of the individual, grouped into \textit{African-American} and \textit{Not African-American}. \\
\hline
Age & Continuous & The age of the individual. \\
\hline
Charge Degree & Categorical & The degree of the charge against the individual. \\
\hline
Priors Count & Continuous & The number of prior convictions or charges. \\
\hline
Risk & Categorical & The risk assessment for recidivism. \\
\hline
Two-Year Recid (target) & Binary & The target variable indicating whether an individual recidivated within two years. \\
\hline
\end{tabular}
\caption{Features in the \textbf{COMPAS} Recidivism Dataset (Preprocessed).}
\label{tab:compas-features}
\end{table*}

\section{Complete Evaluation Results}
\label{appendix:full}

Due to space limitations, we present the complete evaluation results on the \textbf{Adult}, \textbf{German Credit}, and \textbf{COMPAS} datasets in Tables \ref{tab:adult} to \ref{tab:compas}, respectively.

\begin{table*}[!t]
\small
\centering
\begin{tabular}{cc|c|c|c|c|c|c}
  & \multicolumn{3}{c|}{}                                                  & \ACC{}    & \F{}      & \SP{}      & \EoO{}      \\ 
\hline

\multicolumn{1}{c|}{\multirow{18}{*}{\rotatebox{90}{~GPT-3.5-turbo~}  }} & \multirow{3}{*}{\RotText{Zero-Shot~}} & \multirow{3}{*}{}               & \textit{f} & 0.898 \textsubscript{0.001} & 0.711 \textsubscript{ 0.002} & 0.065 \textsubscript{ 0.001} & 0.357 \textsubscript{ 0.000 }  \\
\multicolumn{1}{c|}{}                                    &                             &                                 & \textit{m}   & 0.742 \textsubscript{ 0.002} & 0.727 \textsubscript{ 0.002}& 0.464 \textsubscript{ 0.003} & 0.840 \textsubscript{ 0.004}   \\
\multicolumn{1}{c|}{}                                    &                             &                                 & \textit{d} &  0.157 \textsubscript{ 0.002} & -0.016 \textsubscript{ 0.002} & \textcolor{myred}{-0.399 \textsubscript{ 0.003}} & \textcolor{myred}{-0.483 \textsubscript{ 0.004}}  \\ 
\cdashline{2-8}\noalign{\vskip1.5pt}\cdashline{2-8}
\multicolumn{1}{c|}{}                                    & \multirow{6}{*}{\rotatebox{90}{Few-shot}} & \multirow{3}{*}{{Regular}}        & \textit{f} &  0.899 \textsubscript{ 0.002} & 0.735 \textsubscript{ 0.003} & 0.082 \textsubscript{ 0.002} & 0.429 \textsubscript{ 0.000}  \\
\multicolumn{1}{c|}{}                                    &                             &                                 & \textit{m}   &  0.781 \textsubscript{ 0.003} & 0.749 \textsubscript{ 0.002} & 0.339 \textsubscript{ 0.003} & 0.700 \textsubscript{ 0.003}        \\
\multicolumn{1}{c|}{}                                    &                             &                                 & \textit{d}   &   0.118 \textsubscript{ 0.004} & -0.014 \textsubscript{ 0.004} & -0.257 \textsubscript{ 0.005} \drop{} & -0.271 \textsubscript{0.003}  \drop{} \\ 
\cdashline{3-8}
\multicolumn{1}{c|}{}                                    &                             & \multirow{3}{*}{Label-flipping} & \textit{f} &  0.682 \textsubscript{ 0.004} & 0.590 \textsubscript{ 0.003} & 0.396 \textsubscript{ 0.006} & 0.800 \textsubscript{ 0.013} \\
\multicolumn{1}{c|}{}                                    &                             &                                 & \textit{m}  & 0.614 \textsubscript{ 0.002} & 0.605 \textsubscript{ 0.002} & 0.545 \textsubscript{ 0.001} & 0.763 \textsubscript{ 0.003}  \\
\multicolumn{1}{c|}{}                                    &                             &                                 & \textit{d}   &  0.068 \textsubscript{ 0.004} & -0.015 \textsubscript{ 0.004} & -0.148 \textsubscript{ 0.006} \best{} & 0.037 \textsubscript{ 0.014} \best{}\\ 
\cdashline{2-8}\noalign{\vskip1.5pt}\cdashline{2-8}
\multicolumn{1}{c|}{}                                    & \multirow{9}{*}{\rotatebox{90}{Finetuning}} & \multirow{3}{*}{Regular}        & \textit{f} &  0.915 \textsubscript{ 0.014} & 0.773 \textsubscript{ 0.036} & 0.079 \textsubscript{ 0.002} & 0.476 \textsubscript{ 0.048} \\
\multicolumn{1}{c|}{}                                    &                             &                                 & \textit{m} & 0.799 \textsubscript{ 0.005} & 0.754 \textsubscript{ 0.005} & 0.269 \textsubscript{ 0.036} & 0.613 \textsubscript{ 0.053} \\
\multicolumn{1}{c|}{}                                    &                             &                                 & \textit{d}   &  0.116 \textsubscript{ 0.009} & 0.020 \textsubscript{ 0.039} & -0.190 \textsubscript{ 0.035} \drop{} & -0.137 \textsubscript{ 0.098} \drop{} \\ 
\cdashline{3-8}
\multicolumn{1}{c|}{}                                    &                             & \multirow{3}{*}{Oversampling  } & \textit{f} &  0.913 \textsubscript{0.016} & 0.770 \textsubscript{0.042} &	0.081 \textsubscript{0.004} & 0.476 \textsubscript{0.067}   \\
\multicolumn{1}{c|}{}                                    &                             &                                 & \textit{m}   & 0.813 \textsubscript{0.007} & 0.780 \textsubscript{0.003} & 0.310 \textsubscript{0.038} & 0.702 \textsubscript{0.048} \\
\multicolumn{1}{c|}{}                                    &                             &                                 & \textit{d}   & 0.100 \textsubscript{0.013} & -0.010 \textsubscript{0.041} & -0.229 \textsubscript{0.030} & -0.226 \textsubscript{0.077} \\ 
\cdashline{3-8}
\multicolumn{1}{c|}{}                                    &                             & \multirow{3}{*}{Undersampling}  & \textit{f} &  0.912 \textsubscript{ 0.015} & 0.770 \textsubscript{ 0.046} & 0.086 \textsubscript{ 0.006} & 0.488 \textsubscript{ 0.084} \\
\multicolumn{1}{c|}{}                                    &                             &                                 & \textit{m}   &  0.794 \textsubscript{ 0.006} & 0.751 \textsubscript{ 0.001} & 0.285 \textsubscript{ 0.031} & 0.631 \textsubscript{ 0.044}   \\
\multicolumn{1}{c|}{}                                    &                             &                                 & \textit{d}   &  0.118 \textsubscript{ 0.021} & 0.018 \textsubscript{0.046} & -0.200 \textsubscript{0.025} & -0.143 \textsubscript{ 0.040}\\ 
\hline
\multirow{9}{*}{\rotatebox{90}{\texttt{RF}}}                                      & \multirow{9}{*}{}           & \multirow{3}{*}{Regular}        & \textit{f} &          0.914 \textsubscript{ 0.002} & 0.767 \textsubscript{ 0.006} & 0.075 \textsubscript{ 0.003} & 0.457 \textsubscript{ 0.010}
       \\
  &                             &                                 & \textit{m}   &  0.822 \textsubscript{ 0.005} & 0.783 \textsubscript{ 0.005} & 0.269 \textsubscript{ 0.004} & 0.652 \textsubscript{ 0.004}          \\
  &                             &                                 & \textit{d}   &   0.092 \textsubscript{ 0.004} & -0.015 \textsubscript{ 0.005} & -0.195 \textsubscript{ 0.003} & -0.195 \textsubscript{ 0.012}    \\ 
\cdashline{3-8}
  &                             & \multirow{3}{*}{Oversampling}   & \textit{f} &   0.912 \textsubscript{ 0.006} & 0.770 \textsubscript{ 0.011} & 0.084 \textsubscript{ 0.005} & 0.486 \textsubscript{ 0.012}       \\
  &                             &                                 & \textit{m}   &   0.824 \textsubscript{ 0.002} & 0.785 \textsubscript{ 0.002} & 0.270 \textsubscript{ 0.003} & 0.656 \textsubscript{ 0.006 }     \\
  &                             &                                 & \textit{d}   &      0.087 \textsubscript{ 0.005}  & -0.015 \textsubscript{ 0.01} & -0.185 \textsubscript{ 0.004} & -0.170 \textsubscript{ 0.011}         \\ 
\cdashline{3-8}  
&                             & \multirow{3}{*}{Undersampling}  & \textit{f} &    0.917 \textsubscript{ 0.004} & 0.776 \textsubscript{ 0.011} & 0.075 \textsubscript{ 0.001} & 0.471 \textsubscript{ 0.018}     \\
  &                             &                                 & \textit{m}   &   0.814 \textsubscript{ 0.003} & 0.771 \textsubscript{ 0.004} & 0.263 \textsubscript{ 0.002} & 0.627 \textsubscript{ 0.009}    \\
  &                             &                                 & \textit{d}   &   0.103 \textsubscript{ 0.005} & 0.005 \textsubscript{ 0.011} & -0.187 \textsubscript{ 0.001} & -0.156 \textsubscript{ 0.018}       \\ 
\hline
\multirow{9}{*}{\rotatebox{90}{\texttt{NN}}}                                      & \multirow{9}{*}{}           & \multirow{3}{*}{Regular}        & \textit{f} &    0.917 \textsubscript{ 0.003} & 0.778 \textsubscript{ 0.019} & 0.081 \textsubscript{ 0.016}  & 0.490 \textsubscript{ 0.068}    \\
  &                             &                                 & \textit{m}   &  0.819 \textsubscript{ 0.006} & 0.773 \textsubscript{ 0.015} & 0.250 \textsubscript{ 0.045} & 0.614 \textsubscript{ 0.079}     \\
  &                             &                                 & \textit{d}   &  0.098 \textsubscript{ 0.005} & 0.006 \textsubscript{ 0.009} & -0.169 \textsubscript{ 0.032} & -0.123 \textsubscript{ 0.033}      \\ 
\cdashline{3-8}
  &                             & \multirow{3}{*}{Oversampling}   & \textit{f} &    0.916 \textsubscript{ 0.004} & 0.794 \textsubscript{ 0.013} & 0.100 \textsubscript{ 0.016} & 0.562 \textsubscript{ 0.058}       \\
  &                             &                                 & \textit{m}   &   0.813 \textsubscript{ 0.012} & 0.774 \textsubscript{ 0.008} & 0.286 \textsubscript{ 0.044} & 0.663 \textsubscript{ 0.056}    \\
  &                             &                                 & \textit{d}   &   0.103 \textsubscript{ 0.011} & 0.020 \textsubscript{ 0.018} & -0.186 \textsubscript{ 0.030} & -0.102 \textsubscript{ 0.038}    \\ 
\cdashline{3-8}  
&                             & \multirow{3}{*}{Undersampling}  & \textit{f} &   0.904 \textsubscript{ 0.005} & 0.748 \textsubscript{ 0.014} & 0.084 \textsubscript{ 0.007} & 0.452 \textsubscript{ 0.030}       \\
  &                             &                                 & \textit{m}   &  0.813 \textsubscript{ 0.006} & 0.774 \textsubscript{ 0.005} & 0.283 \textsubscript{ 0.023} & 0.659 \textsubscript{ 0.031}       \\
  &                             &                                 & \textit{d}   &    0.090 \textsubscript{ 0.006} & -0.026 \textsubscript{ 0.014} & -0.199 \textsubscript{ 0.018} & -0.206 \textsubscript{ 0.031}       \\ 
\hline
\end{tabular}
\caption{\textbf{Fairness evaluation for Adult dataset}. This table depicts the evaluation of accuracy (\ACC{}), \F{} score (\F{}), statistical parity (\SP{}), and equality of opportunity (\EoO{}) metrics for the subgroup - \textit{female} (\textit{f}) and \textit{male} (\textit{m}) as well as the difference (\textit{d}) between them. We list the protected group first. The significant fairness disparities are highlighted in \textcolor{myred}{red}. Both in-context learning and finetuning can lead to bias reduction (indicated by 
\drop{}), and label-flipped in-context learning can further minimize bias (indicated by \best{}).}
\label{tab:adult}
\vspace{-8pt}
\end{table*}

\begin{table*}[!t]
\small
\centering
\begin{tabular}{cc|c|c|c|c|c|c}
  & \multicolumn{3}{c|}{}                                                  & \ACC{}    & \F{}      & \SP{}      & \EoO{}      \\ 
\hline
\multicolumn{1}{c|}{\multirow{18}{*}{\rotatebox{90}{~GPT-3.5-turbo~}  }} & \multirow{3}{*}{\RotText{Zero-Shot~}} & \multirow{3}{*}{}               & \textit{f} &  0.471 \textsubscript{ 0.011} & 0.359 \textsubscript{ 0.021} & 0.980 \textsubscript{ 0.011} & 1.000 \textsubscript{ 0.000}  \\
\multicolumn{1}{c|}{}                                    &                             &                                 & \textit{m}   &  0.556 \textsubscript{ 0.000} & 0.357 \textsubscript{ 0.000} & 0.984 \textsubscript{ 0.000} & 0.972 \textsubscript{ 0.000}  \\
\multicolumn{1}{c|}{}                                    &                             &                                 & \textit{d}   & -0.084 \textsubscript{ 0.011} & 0.002 \textsubscript{ 0.021} & -0.004 \textsubscript{ 0.011} & 0.028 \textsubscript{ 0.000} \\ 
\cdashline{2-8}\noalign{\vskip1.5pt}\cdashline{2-8}
\multicolumn{1}{c|}{}                                    & \multirow{6}{*}{\rotatebox{90}{Few-shot}} & \multirow{3}{*}{{Regular}}        & \textit{f} & 0.610 \textsubscript{ 0.013} & 0.593 \textsubscript{ 0.013}  & 0.348 \textsubscript{ 0.027} & 0.453 \textsubscript{ 0.029}  \\
\multicolumn{1}{c|}{}                                    &                             &                                 & \textit{m}   &   0.606 \textsubscript{ 0.007} & 0.603 \textsubscript{ 0.008} & 0.337 \textsubscript{ 0.007} & 0.450 \textsubscript{ 0.012}     \\
\multicolumn{1}{c|}{}                                    &                             &                                 & \textit{d}   &  0.003 \textsubscript{ 0.012} & -0.010 \textsubscript{ 0.011} & 0.011 \textsubscript{ 0.027} & 0.003 \textsubscript{ 0.026}  \\ 
\cdashline{3-8}
\multicolumn{1}{c|}{}                                    &                             & \multirow{3}{*}{Label-flipping} & \textit{f} & 0.614 \textsubscript{ 0.011} & 0.606 \textsubscript{ 0.012} & 0.695 \textsubscript{ 0.011} & 0.842 \textsubscript{ 0.000}   \\
\multicolumn{1}{c|}{}                                    &                             &                                 & \textit{m}   &  0.559 \textsubscript{ 0.013} & 0.538 \textsubscript{ 0.011} & 0.638 \textsubscript{ 0.013} & 0.672 \textsubscript{ 0.023}  \\
\multicolumn{1}{c|}{}                                    &                             &                                 & \textit{d}   &   0.056 \textsubscript{ 0.021} & 0.067 \textsubscript{ 0.021} & 0.057 \textsubscript{ 0.012} & 0.170 \textsubscript{ 0.023} \\ 
\cdashline{2-8}\noalign{\vskip1.5pt}\cdashline{2-8}
\multicolumn{1}{c|}{}                                    & \multirow{9}{*}{\rotatebox{90}{Finetuning}} & \multirow{3}{*}{Regular}        & \textit{f} & 0.571 \textsubscript{ 0.067} & 0.567 \textsubscript{ 0.062} & 0.619 \textsubscript{ 0.101} & 0.711 \textsubscript{ 0.186} \\
\multicolumn{1}{c|}{}                                    &                             &                                 & \textit{m} &  0.548 \textsubscript{ 0.011} & 0.539 \textsubscript{ 0.023} & 0.532 \textsubscript{ 0.123} & 0.569 \textsubscript{ 0.098} \\
\multicolumn{1}{c|}{}                                    &                             &                                 & \textit{d}   &  0.024 \textsubscript{ 0.079} & 0.029 \textsubscript{ 0.085} & 0.087 \textsubscript{ 0.022} & 0.141 \textsubscript{ 0.088}  \\ 
\cdashline{3-8}
\multicolumn{1}{c|}{}                                    &                             & \multirow{3}{*}{Oversampling  } & \textit{f} &   0.536 \textsubscript{ 0.017} & 0.532 \textsubscript{ 0.012} & 0.607 \textsubscript{ 0.084} & 0.658 \textsubscript{ 0.112}  \\
\multicolumn{1}{c|}{}                                    &                             &                                 & \textit{m} &  0.532 \textsubscript{ 0.011} & 0.523 \textsubscript{ 0.020} & 0.548 \textsubscript{ 0.079} & 0.569 \textsubscript{ 0.059}\\
\multicolumn{1}{c|}{}                                    &                             &                                 & \textit{d}   &   0.004 \textsubscript{ 0.028} & 0.009 \textsubscript{ 0.033} & 0.060 \textsubscript{ 0.006} & 0.088 \textsubscript{ 0.053}   \\ 
\cdashline{3-8}
\multicolumn{1}{c|}{}                                    &                             & \multirow{3}{*}{Undersampling}  & \textit{f} & 0.548 \textsubscript{ 0.034} & 0.547 \textsubscript{ 0.033} & 0.571 \textsubscript{ 0.034} & 0.632 \textsubscript{ 0.074} \\
\multicolumn{1}{c|}{}                                    &                             &                                 & \textit{m}   & 0.556 \textsubscript{ 0.000} & 0.555 \textsubscript{ 0.000} & 0.444 \textsubscript{ 0.000} & 0.500 \textsubscript{ 0.000} \\
\multicolumn{1}{c|}{}                                    &                             &                                 & \textit{d}   &  -0.008 \textsubscript{ 0.034} & -0.008 \textsubscript{ 0.033} & 0.127 \textsubscript{ 0.034} & 0.132 \textsubscript{ 0.074} \\ 
\hline
\multirow{9}{*}{\rotatebox{90}{\texttt{RF}}}                                      & \multirow{9}{*}{}           & \multirow{3}{*}{Regular}        & \textit{f} &     0.581 \textsubscript{ 0.024} & 0.580 \textsubscript{ 0.025} & 0.519 \textsubscript{ 0.028} & 0.611 \textsubscript{ 0.054}         \\
  &                             &                                 & \textit{m}   &  0.600 \textsubscript{ 0.019} & 0.588 \textsubscript{ 0.020} & 0.597 \textsubscript{ 0.022} & 0.672 \textsubscript{ 0.021}       \\
  &                             &                                 & \textit{d}   &   -0.019 \textsubscript{ 0.016} & -0.008 \textsubscript{ 0.016} & -0.078 \textsubscript{ 0.044} & -0.062 \textsubscript{ 0.061}       \\ 
\cdashline{3-8}
  &                             & \multirow{3}{*}{Oversampling}   & \textit{f} &    0.576 \textsubscript{ 0.018} & 0.575 \textsubscript{ 0.018} & 0.505 \textsubscript{ 0.018} & 0.589 \textsubscript{ 0.021}       \\
  &                             &                                 & \textit{m}   &  0.568 \textsubscript{ 0.032} & 0.552 \textsubscript{ 0.034} & 0.616 \textsubscript{ 0.025} & 0.661 \textsubscript{ 0.037}    \\
  &                             &                                 & \textit{d}   &   0.008 \textsubscript{ 0.034} & 0.023 \textsubscript{ 0.035} & -0.111 \textsubscript{ 0.013} & -0.072 \textsubscript{ 0.041}         \\ 
\cdashline{3-8}  
&                             & \multirow{3}{*}{Undersampling}  & \textit{f} &   0.586 \textsubscript{ 0.024} & 0.585 \textsubscript{ 0.024}   & 0.533 \textsubscript{ 0.024} & 0.632 \textsubscript{ 0.047}     \\
  &                             &                                 & \textit{m}   &   0.575 \textsubscript{ 0.031} & 0.555 \textsubscript{ 0.037} & 0.635 \textsubscript{ 0.033} & 0.683 \textsubscript{ 0.022}        \\
  &                             &                                 & \textit{d}   &   0.011 \textsubscript{ 0.024} & 0.031 \textsubscript{ 0.031} & -0.102 \textsubscript{ 0.041} & -0.052 \textsubscript{ 0.039}        \\ 
\hline
\multirow{9}{*}{\rotatebox{90}{\texttt{NN}}}                                      & \multirow{9}{*}{}           & \multirow{3}{*}{Regular}        & \textit{f} &   0.533 \textsubscript{ 0.024} & 0.533 \textsubscript{ 0.024} & 0.519 \textsubscript{ 0.028} & 0.558 \textsubscript{ 0.026}     \\
  &                             &                                 & \textit{m}   &    0.556 \textsubscript{ 0.017} & 0.544 \textsubscript{ 0.017} & 0.584 \textsubscript{ 0.012} & 0.622 \textsubscript{ 0.022}         \\
  &                             &                                 & \textit{d}   &   -0.022 \textsubscript{ 0.037} & -0.012 \textsubscript{ 0.036} & -0.065 \textsubscript{ 0.031} & -0.064 \textsubscript{ 0.026}      \\ 
\cdashline{3-8}
  &                             & \multirow{3}{*}{Oversampling}   & \textit{f} &    0.548 \textsubscript{ 0.040} & 0.547 \textsubscript{ 0.040} & 0.552 \textsubscript{ 0.028} & 0.611 \textsubscript{ 0.026 }        \\
  &                             &                                 & \textit{m}   &    0.562 \textsubscript{ 0.026} & 0.547 \textsubscript{ 0.024} & 0.603 \textsubscript{ 0.048} & 0.644 \textsubscript{ 0.057}    \\
  &                             &                                 & \textit{d}   &  -0.014 \textsubscript{ 0.037} & 0.000 \textsubscript{ 0.035} & -0.051 \textsubscript{ 0.061} & -0.034 \textsubscript{ 0.065}      \\ 
\cdashline{3-8}  
&                             & \multirow{3}{*}{Undersampling}  & \textit{f} &   0.529 \textsubscript{ 0.049} & 0.524 \textsubscript{ 0.047} & 0.467 \textsubscript{ 0.051} & 0.495 \textsubscript{ 0.042}    \\
  &                             &                                 & \textit{m}   &   0.495 \textsubscript{ 0.025} & 0.490 \textsubscript{ 0.023} & 0.524 \textsubscript{ 0.047} & 0.517 \textsubscript{ 0.054}       \\
  &                             &                                 & \textit{d}   &  0.033 \textsubscript{ 0.063} & 0.035 \textsubscript{ 0.059} &-0.057 \textsubscript{ 0.033} & -0.022 \textsubscript{ 0.061}        \\ 
\hline
\end{tabular}
\caption{\textbf{Fairness evaluation for German Credit dataset}. This table depicts the evaluation of accuracy (\ACC{}), \F{} score (\F{}), statistical parity (\SP{}), and equality of opportunity (\EoO{}) metrics for the subgroup - \textit{female} (\textit{f}) and \textit{male} (\textit{m}) as well as the difference (\textit{d}) between them. }
\label{tab:germancredit}
\vspace{-8pt}
\end{table*}

\begin{table*}[!t]
\small
\centering
\begin{tabular}{cc|c|c|c|c|c|c}
  & \multicolumn{3}{c|}{}                                                  & \ACC{}    & \F{}      & \SP{}      & \EoO{}      \\ 
\hline
\multicolumn{1}{c|}{\multirow{18}{*}{\rotatebox{90}{~GPT-3.5-turbo~}  }} & \multirow{3}{*}{\RotText{Zero-Shot~}} & \multirow{3}{*}{}               & \textit{AA} & 0.657 \textsubscript{ 0.005} & 0.656 \textsubscript{ 0.004} & 0.395 \textsubscript{ 0.001} & 0.560 \textsubscript{ 0.002}   \\
\multicolumn{1}{c|}{}                                    &                             &                                 & \textit{nAA}   &  0.663 \textsubscript{ 0.002} & 0.588 \textsubscript{ 0.003} & 0.817 \textsubscript{ 0.002} & 0.893 \textsubscript{ 0.001} \\
\multicolumn{1}{c|}{}                                    &                             &                                 & \textit{d}   &  -0.006 \textsubscript{ 0.005} & 0.068 \textsubscript{ 0.006} & \textcolor{myred}{-0.423 \textsubscript{ 0.003}} & \textcolor{myred}{-0.334 \textsubscript{ 0.002}} \\ 
\cdashline{2-8}\noalign{\vskip1.5pt}\cdashline{2-8}
\multicolumn{1}{c|}{}                                    & \multirow{6}{*}{\rotatebox{90}{Few-shot}} & \multirow{3}{*}{{Regular}}        & \textit{AA} & 0.633 \textsubscript{ 0.002} & 0.626 \textsubscript{ 0.002} & 0.362 \textsubscript{ 0.003} & 0.495 \textsubscript{ 0.004}   \\
\multicolumn{1}{c|}{}                                    &                             &                                 & \textit{nAA}   & 0.642 \textsubscript{ 0.001} & 0.623 \textsubscript{ 0.002} & 0.614 \textsubscript{ 0.002} & 0.709 \textsubscript{ 0.002}     \\
\multicolumn{1}{c|}{}                                    &                             &                                 & \textit{d}   &  -0.008 \textsubscript{ 0.003} & 0.003 \textsubscript{ 0.003} & -0.252 \textsubscript{ 0.003} \drop{} & -0.214 \textsubscript{ 0.005} \drop{} \\ 
\cdashline{3-8}
\multicolumn{1}{c|}{}                                    &                             & \multirow{3}{*}{Label-flipping} & \textit{AA} & 0.482 \textsubscript{ 0.004} & 0.482 \textsubscript{ 0.004} & 0.499 \textsubscript{ 0.003} & 0.481 \textsubscript{ 0.004} \\
\multicolumn{1}{c|}{}                                    &                             &                                 & \textit{nAA}   & 0.412 \textsubscript{ 0.003} & 0.408 \textsubscript{ 0.003} & 0.471 \textsubscript{ 0.002} & 0.404 \textsubscript{ 0.003} \\
\multicolumn{1}{c|}{}                                    &                             &                                 & \textit{d}   &  0.070 \textsubscript{ 0.005} & 0.074 \textsubscript{ 0.005} & 0.028 \textsubscript{ 0.005} \best{} & 0.077 \textsubscript{ 0.007} \best{}   \\ 
\cdashline{2-8}\noalign{\vskip1.5pt}\cdashline{2-8}
\multicolumn{1}{c|}{}                                    & \multirow{9}{*}{\rotatebox{90}{Finetuning}} & \multirow{3}{*}{Regular}        & \textit{AA} &  0.611 \textsubscript{ 0.016} & 0.610 \textsubscript{ 0.016} & 0.464 \textsubscript{ 0.031} & 0.576 \textsubscript{ 0.034}  \\
\multicolumn{1}{c|}{}                                    &                             &                                 & \textit{nAA} &  0.616 \textsubscript{ 0.013} & 0.586 \textsubscript{ 0.016} & 0.657 \textsubscript{ 0.032} & 0.724 \textsubscript{ 0.029} \\
 \multicolumn{1}{c|}{}                                    &                             &                                 & \textit{d}   &  -0.005 \textsubscript{ 0.017} & 0.024 \textsubscript{ 0.024} & -0.193 \textsubscript{ 0.030} \drop{} & -0.148 \textsubscript{ 0.027} \drop{} \\ 
\cdashline{3-8}
\multicolumn{1}{c|}{}                                    &                             & \multirow{3}{*}{Oversampling  } & \textit{AA} &  0.609 \textsubscript{ 0.007} & 0.608 \textsubscript{ 0.007} & 0.494 \textsubscript{ 0.071} & 0.605 \textsubscript{ 0.066} \\
\multicolumn{1}{c|}{}                                    &                             &                                 & \textit{nAA} &0.625 \textsubscript{ 0.020} & 0.583 \textsubscript{ 0.024} & 0.706 \textsubscript{ 0.037} & 0.771 \textsubscript{ 0.036}
\\
\multicolumn{1}{c|}{}                                    &                             &                                 & \textit{d}   &  -0.016 \textsubscript{ 0.016} & 0.025 \textsubscript{ 0.018} & -0.212 \textsubscript{ 0.037} & -0.166 \textsubscript{ 0.046}  \\ 
\cdashline{3-8}
\multicolumn{1}{c|}{}                                    &                             & \multirow{3}{*}{Undersampling}  & \textit{AA} & 0.591 \textsubscript{0.010} & 0.591 \textsubscript{0.012} & 0.513 \textsubscript{0.053} & 0.605 \textsubscript{0.047}\\
\multicolumn{1}{c|}{}                                    &                             &                                 & \textit{nAA}   & 0.641 \textsubscript{0.008} & 0.612 \textsubscript{0.009} & 0.663 \textsubscript{0.035} & 0.749 \textsubscript{0.037}\\
\multicolumn{1}{c|}{}                                    &                             &                                 & \textit{d}   &  -0.050 \textsubscript{0.016} & -0.021 \textsubscript{0.022} & -0.150 \textsubscript{0.033} & -0.144 \textsubscript{0.039}  \\ 
\hline
\multirow{9}{*}{\rotatebox{90}{\texttt{RF}}}                                      & \multirow{9}{*}{}           & \multirow{3}{*}{Regular}        & \textit{AA} &       0.662 \textsubscript{ 0.004} & 0.662 \textsubscript{ 0.004} & 0.496 \textsubscript{ 0.006} & 0.660 \textsubscript{ 0.007}       \\
  &                             &                                 & \textit{nAA}   &     0.671 \textsubscript{ 0.004} & 0.617 \textsubscript{ 0.002} & 0.767 \textsubscript{ 0.008} & 0.859 \textsubscript{ 0.009}        \\
  &                             &                                 & \textit{d}   &     -0.009 \textsubscript{ 0.007} & 0.045 \textsubscript{ 0.005} & -0.271 \textsubscript{ 0.011} & -0.199 \textsubscript{ 0.014}     \\ 
\cdashline{3-8}
  &                             & \multirow{3}{*}{Oversampling}   & \textit{AA} &     0.660 \textsubscript{ 0.005} & 0.660 \textsubscript{ 0.005} & 0.493 \textsubscript{ 0.010} & 0.655 \textsubscript{ 0.013}        \\
  &                             &                                 & \textit{nAA}   &     0.671 \textsubscript{ 0.002} & 0.624 \textsubscript{ 0.002} & 0.743 \textsubscript{ 0.003} & 0.839 \textsubscript{ 0.004}         \\
  &                             &                                 & \textit{d}   &   -0.010 \textsubscript{ 0.006} & 0.037 \textsubscript{ 0.006} & -0.250 \textsubscript{ 0.012} & -0.184 \textsubscript{ 0.016}         \\ 
\cdashline{3-8}  
&                             & \multirow{3}{*}{Undersampling}  & \textit{AA} &    0.648 \textsubscript{ 0.002} & 0.647 \textsubscript{ 0.002} & 0.491 \textsubscript{ 0.004} & 0.639 \textsubscript{ 0.004}        \\
  &                             &                                 & \textit{nAA}   &  0.667 \textsubscript{ 0.005} & 0.614 \textsubscript{ 0.007} & 0.761 \textsubscript{ 0.006} & 0.851 \textsubscript{ 0.006}        \\
  &                             &                                 & \textit{d}   &   -0.020 \textsubscript{ 0.007} & 0.033 \textsubscript{ 0.008} & -0.270 \textsubscript{ 0.009} & -0.211 \textsubscript{ 0.008}        \\ 
\hline
\multirow{9}{*}{\rotatebox{90}{\texttt{NN}}}                                      & \multirow{9}{*}{}           & \multirow{3}{*}{Regular}        & \textit{AA} &     0.666 \textsubscript{ 0.003} & 0.665 \textsubscript{ 0.002} & 0.462 \textsubscript{ 0.034} & 0.630 \textsubscript{ 0.034}        \\
  &                             &                                 & \textit{nAA}   &     0.662 \textsubscript{ 0.003} & 0.613 \textsubscript{ 0.006} & 0.742 \textsubscript{ 0.019} & 0.831 \textsubscript{ 0.017}         \\
  &                             &                                 & \textit{d}   &   0.005 \textsubscript{ 0.006} & 0.052 \textsubscript{ 0.007} & -0.280 \textsubscript{ 0.019} & -0.201 \textsubscript{ 0.018}        \\ 
\cdashline{3-8}
  &                             & \multirow{3}{*}{Oversampling}   & \textit{AA} &    0.656 \textsubscript{ 0.001} & 0.653 \textsubscript{ 0.012} & 0.507 \textsubscript{ 0.090} & 0.665 \textsubscript{ 0.101}     \\
  &                             &                                 & \textit{nAA}   &   0.643 \textsubscript{ 0.013} & 0.580 \textsubscript{ 0.034} & 0.757 \textsubscript{ 0.107} & 0.828 \textsubscript{ 0.091}         \\
  &                             &                                 & \textit{d}   &    0.013 \textsubscript{ 0.014} & 0.073 \textsubscript{ 0.043} & -0.249 \textsubscript{ 0.049} & -0.163 \textsubscript{ 0.046}      \\ 
\cdashline{3-8}  
&                             & \multirow{3}{*}{Undersampling}  & \textit{AA} &     0.660 \textsubscript{ 0.019} & 0.657 \textsubscript{ 0.023} & 0.477 \textsubscript{ 0.078} & 0.638 \textsubscript{ 0.097}      \\
  &                             &                                 & \textit{nAA}   &    0.657 \textsubscript{ 0.013} & 0.602 \textsubscript{ 0.026} & 0.757 \textsubscript{ 0.051} & 0.839 \textsubscript{ 0.040}       \\
  &                             &                                 & \textit{d}   &    0.003 \textsubscript{ 0.024} & 0.055 \textsubscript{ 0.043} & -0.280 \textsubscript{ 0.041} & -0.202 \textsubscript{ 0.064}      \\ 
\hline
\end{tabular}
\caption{\textbf{Fairness evaluation for COMPAS dataset} for the subgroup - \textit{African American} \textit{(AA)}, and \textit{Non African American} (\textit{nAA}) as well as the difference (\textit{d}). The significant fairness disparities are highlighted in \textcolor{myred}{red}. Both in-context learning and finetuning can lead to bias reduction (indicated by 
\drop{}), and label-flipped in-context learning can further minimize bias (indicated by \best{}).}
\label{tab:compas}
\vspace{-8pt}
\end{table*}

\section{RF and NN Hyperparameters}
\label{appendix:rf_nn_hp}
For \texttt{RF}, we fix the number of trees to 100 for all datasets as well as models. For \texttt{NN}, we use a 3 hidden-layered network with hyperparameters described in Table \ref{tab:hyper_nn}.

\begin{table}[!ht]
\resizebox{0.49 \textwidth}{!}{
\centering
\begin{tabular}{|c|c|c|c|c|c|c|} 
\hline
              & \textbf{h1} & \textbf{h2} & \textbf{h3} & \textbf{lr} & \textbf{batch size} & \textbf{epochs}  \\ 
\hline\hline
\textbf{Adult}         & 16          & 64          & 16          & 0.07        & 128                 & 300              \\ 
\hline
\textbf{German Credit} & 64          & 64          & 32          & 0.07        & 128                 & 300              \\ 
\hline
\textbf{COMPAS}        & 64          & 128         & 64          & 0.09        & 128                 & 300              \\
\hline
\end{tabular}}
\caption{Hyperparameters for all datasets for a 3-layer neural network, where h1, h2, and h3 represent the number of neurons in first, second, and third hidden layers respectively, lr represents the learning rate and is followed by the batch size and the number of epochs the models are trained for.}
\label{tab:hyper_nn}
\end{table}

\section{Prompt Templates}
\label{appendix:prompts}
Beyond the Adult dataset, we provide the serialization and prompt templates utilized in our work for the German Credit and COMPAS datasets here.
As before, text in \textcolor{myblue}{blue} denotes the task description, text in \textcolor{mygreen}{green} indicates optional few-shot examples (only used in in-context learning setting), and text in \textcolor{myred}{red} is the test example.

\subsection{German Credit}
\vspace{0.2cm}

\begin{tcolorbox}[prompt, title = {{Prompt 2. Prompt Template for \texttt{German Credit} Dataset.}}]
{
\vspace{-0.2cm}
\begin{lstlisting}[
    breaklines=true,
    postbreak=\mbox{\hspace{-0.7cm}},
    basicstyle=\color{myblue}\ttfamily\footnotesize,
    columns=fullflexible,
    escapeinside={(*}{*)}
]
Predict the credit risk of a German bank customer based on their profile data. Answer with one of the following: bad | good.
\end{lstlisting}
}

{
\vspace{-0.4cm}
\begin{lstlisting}[
    breaklines=true,
    postbreak=\mbox{\hspace{-0.7cm}},
    basicstyle=\color{mygreen}\ttfamily\footnotesize,
    columns=fullflexible,
    escapeinside={(*}{*)}
]
Example 1 -
Age: 25 - 45
Sex: female
Job: highly skilled
Housing: rent
Saving accounts: little
Checking account: little
Credit amount: 2606
Duration: 21
Purpose: radio/TV
Answer: good
...
\end{lstlisting}
}

{
\vspace{-0.3cm}
\begin{lstlisting}[
    breaklines=true,
    postbreak=\mbox{\hspace{-0.7cm}},
    basicstyle=\color{myred}\ttfamily\footnotesize,
    columns=fullflexible,
    escapeinside={(*}{*)}
]
Age: 25 - 45
Sex: male
Job: skilled
Housing: own
Saving accounts: little
Checking account: little
Credit amount: 1345
Duration: 18
Purpose: radio/TV
Answer:
\end{lstlisting}
}
\vspace{-0.25cm}
\end{tcolorbox}

\subsection{COMPAS}
\vspace{0.2cm}

\begin{tcolorbox}[prompt, title = {{Prompt 3. Prompt Template for \texttt{COMPAS} Dataset.}}]
{
\vspace{-0.2cm}
\begin{lstlisting}[
    breaklines=true,
    postbreak=\mbox{\hspace{-0.7cm}},
    basicstyle=\color{myblue}\ttfamily\footnotesize,
    columns=fullflexible,
    escapeinside={(*}{*)}
]
Predict whether an individual will recidivate
with in two years based on demographic and criminal history. Answer with one of the following: Did Not Reoffend | Reoffended.
\end{lstlisting}
}
{
\vspace{-0.4cm}
\begin{lstlisting}[
    breaklines=true,
    postbreak=\mbox{\hspace{-0.7cm}},
    basicstyle=\color{mygreen}\ttfamily\footnotesize,
    columns=fullflexible,
    escapeinside={(*}{*)}
]
Example 1 -
sex: Male
race: African-American
age cat: 25 - 45
c charge degree: F
priors count: 0
risk: Low
Answer: Did Not Reoffend
...
\end{lstlisting}
}

{
\vspace{-0.3cm}
\begin{lstlisting}[
    breaklines=true,
    postbreak=\mbox{\hspace{-0.7cm}},
    basicstyle=\color{myred}\ttfamily\footnotesize,
    columns=fullflexible,
    escapeinside={(*}{*)}
]
sex: Male
race: African-American
age cat: 25 - 45
c charge degree: M
priors count: 13
risk: High
Answer:
\end{lstlisting}
}
\vspace{-0.25cm}
\end{tcolorbox}

\end{document}